\documentclass[manuscript,screen]{acmart}
\usepackage{multirow}
\usepackage{tipa}

\AtBeginDocument{%
  }

\copyrightyear{2026}
\acmYear{2026}
\setcopyright{cc}
\setcctype{by}
\acmConference[MM '26]{Proceedings of the 34th ACM International Conference on Multimedia}{November 10--14, 2026}{Rio de Janeiro, Brazil}
\acmBooktitle{Proceedings of the 34th ACM International Conference on Multimedia (MM '26), November 10--14, 2026, Rio de Janeiro, Brazil}
\acmDOI{10.1145/3767308.3835096}
\acmISBN{979-8-4007-2213-4/2026/11}

\begin{document}

\title{PD-GS: Phoneme-Driven 3DGS for Audio-Driven Talking Heads}


\author{Ao Fu}
\orcid{0009-0006-9933-9240}
\email{220232248@seu.edu.cn}
\affiliation{%
  \department{School of Computer Science and Engineering}
  \institution{Southeast University}
  \city{Nanjing}
  \country{China}}
\affiliation{%
  \department{Key Laboratory of New Generation Artificial Intelligence
    Technology and Its Interdisciplinary Applications}
  \institution{Ministry of Education}
  \city{Nanjing}
  \country{China}}

\author{Yi Zhou}
\orcid{0000-0003-3021-3229}
\authornote{Corresponding author.}
\email{yizhou.szcn@gmail.com}
\affiliation{%
  \department{School of Computer Science and Engineering}
  \institution{Southeast University}
  \city{Nanjing}
  \country{China}}
\affiliation{%
  \department{Key Laboratory of New Generation Artificial Intelligence
    Technology and Its Interdisciplinary Applications}
  \institution{Ministry of Education}
  \city{Nanjing}
  \country{China}}
\renewcommand{\shortauthors}{Fu and Zhou}

\begin{abstract}
3D Gaussian Splatting (3DGS) enables fast, photorealistic talking-head rendering, yet accurate lip articulation remains elusive: mouth motion is often over-smoothed and may violate hard articulatory constraints such as bilabial closures, producing the notorious ``leaky mouth'' artifact. A key difficulty is that brief, discrete articulatory events are inferred from a continuous acoustic embedding under a regression objective, which biases predictions toward averaged mouth configurations. While modern self-supervised speech encoders provide rich prosodic and phonetic cues, they do not provide an explicit, frame-aligned linguistic target that reliably disambiguates closure-level events. We propose \textbf{Phoneme-Driven Gaussian Splatting (PD-GS)}, which augments a 3DGS talker with time-aligned phoneme tokens obtained from an automatic ASR and forced-alignment pipeline. Our core component, the \textbf{Linguistic Fusion Module (LFM)}, adaptively fuses continuous audio context with discrete phoneme embeddings through a learned gate, allowing the model to preserve smooth audio-driven dynamics while strengthening phoneme guidance on articulation-critical segments. PD-GS is trained purely from monocular video using image reconstruction and lip landmark supervision. On HDTF, PD-GS achieves the best lip geometry among the compared baselines (LMD 2.66) and qualitatively reduces closure violations in challenging phoneme sequences, yielding more linguistically faithful neural avatars.
\end{abstract}



\begin{teaserfigure}
    \centering
    \includegraphics[width=0.95\linewidth]{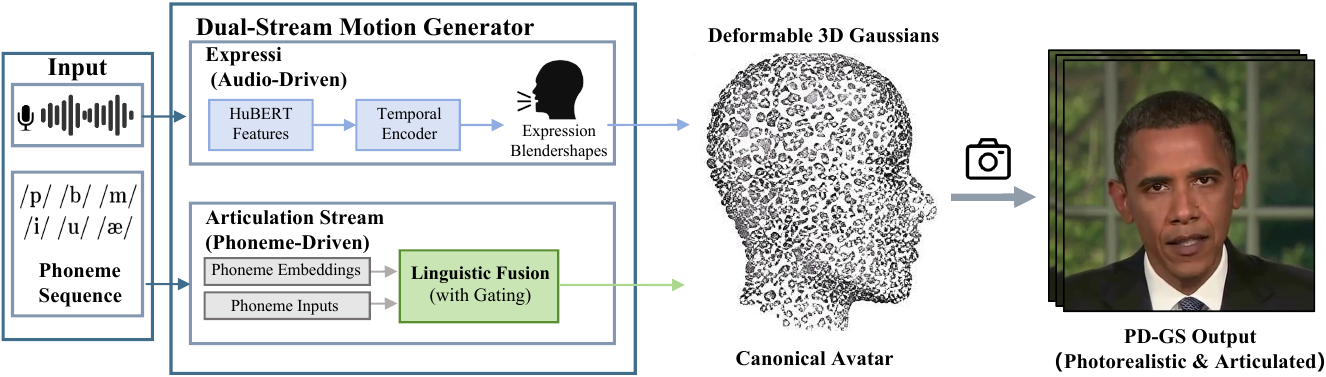} 
    \caption{
    \textbf{Dual-Stream Motion Generator.}
    Two complementary streams drive a 3D Gaussian avatar:
    a \textbf{prosody stream} predicts expression blendshape weights from HuBERT audio features, and
    an \textbf{articulation stream} fuses time-aligned phonemes through a \textbf{Linguistic Fusion Module (LFM)} with a learned gating network to produce precise lip motions,
    enabling photorealistic rendering with linguistically faithful articulation.}
    \Description{A two-branch architecture diagram. HuBERT audio features feed
    an expression branch for global facial motion and an articulation branch
    that fuses audio features with aligned phoneme embeddings. The predicted
    expression and lip deformations are combined to animate and render a
    canonical 3D Gaussian avatar.}
    \label{fig:1}
\end{teaserfigure}


\maketitle

\section{Introduction}
The advent of 3D Gaussian Splatting (3DGS)~\cite{kerbl20233d} has substantially alleviated the rendering bottleneck for audio-driven talking heads, enabling photorealistic avatars with impressive speed and visual fidelity~\cite{gaussiantalker}. However, even strong recent 3DGS-based talkers still exhibit a conspicuous articulation failure: mouth shapes are often over-smoothed and ``muffled,'' and the lips frequently fail to fully seal during bilabial closures, producing the well-known ``leaky mouth'' artifact. This indicates that the remaining bottleneck is no longer rendering quality, but rather the \emph{physical and linguistic correctness} of facial motion. Beyond academic benchmarks, accurate audio-driven facial motion is increasingly demanded in real-world applications such as telepresence, virtual assistants, digital humans, film dubbing, and VR/AR communication. In these scenarios, viewers are especially sensitive to articulation errors: even when the overall rendering is photorealistic, subtle mis-closures and over-smoothed lip dynamics can immediately break the illusion of speaking, degrade intelligibility, and reduce perceived naturalness. Therefore, articulation faithfulness is essential for deploying high-fidelity avatars. From a multimedia perspective, this problem is inherently multimodal: a realistic talking head must consistently align acoustic evidence, linguistic structure, facial motion, and photorealistic visual rendering over time.

We contend that the prevailing end-to-end mapping from \emph{continuous} audio features to facial deformation remains under-constrained for \emph{discrete} articulatory events, especially closures and rapid transitions. Throughout the paper, we use /\(\cdot\)/ to denote phonemes (abstract linguistic units) and [\(\cdot\)] to denote phone-level articulation examples (surface realizations). Speech science shows that each phonetic unit corresponds to a characteristic facial activation pattern modulated by context---a phenomenon known as \textbf{co-articulation}~\cite{leanderson1971electromyographic, lucero2012speech}. For example, the articulatory pattern for /p/ differs markedly depending on whether it follows /i/ (a spread-lip vowel) or /u/ (a rounded-lip vowel). While modern systems often use deep self-supervised audio encoders (e.g., HuBERT~\cite{hsu2021hubert}) whose representations implicitly contain phonetic cues, they typically lack an \emph{explicit, time-aligned, discrete linguistic target} that can enforce closure-level constraints when the audio evidence is ambiguous or smeared by context. As a result, regressors tend to predict ``averaged'' mouth shapes and under-shoot closure amplitudes or durations, leading to the characteristic ``leaky mouth'' artifact.

In this paper, we present \textbf{Phoneme-Driven Gaussian Splatting (PD-GS)}, a multimodal framework that restores explicit linguistic grounding for high-fidelity 3DGS-based talking heads. PD-GS explicitly combines complementary cues from continuous acoustic features and discrete, time-aligned phoneme tokens, allowing the model to preserve smooth audio-driven dynamics while enforcing precise linguistic constraints for articulation. Phonemes have been used in prior animation systems, e.g., as supervision or regularization in mesh- or parameter-driven pipelines~\cite{cudeiro2019capture, imitator}. Our focus is complementary: we integrate phoneme-level control into a photorealistic 3DGS pipeline \emph{as an auxiliary input at inference time} via a lightweight \textbf{Linguistic Fusion Module (LFM)}. Specifically, we extract a unified representation using a HuBERT encoder, and then drive two decoupled motion streams: an \textbf{Expression Module} for prosody-driven motion and an \textbf{Articulation Module} equipped with the LFM. The LFM takes an auxiliary stream of discrete, time-aligned phonemes---obtained via a fully automated ASR and forced-alignment pipeline~\cite{radford2023robust, mcauliffe2017montreal}---and uses a learned gate to adaptively modulate the contribution of phoneme and audio cues. In this paper, we interpret the gate as learning the compatibility between local acoustic context and aligned phoneme tokens, rather than as receiving an explicit external confidence signal. This design preserves rich continuous audio context while allowing discrete linguistic targets to contribute more strongly on articulation-critical segments such as bilabial closures and strong rounding.

Our contributions are threefold:
\begin{itemize}
    \item We present PD-GS, a dual-stream 3DGS talking-head framework that injects aligned phoneme tokens through a Linguistic Fusion Module (LFM) to improve articulation.
    \item PD-GS achieves the best LMD among the compared methods on HDTF while remaining competitive in perceptual quality, synchronization, and model-side latency.
    \item Additional analyses, including user study, gate visualization, and cross-dataset evaluation, show that phoneme grounding improves both interpretability and generalization.
\end{itemize}

\section{Related Work}

\subsection{Audio-Driven Talking Head Synthesis}
Audio-driven talking head synthesis has evolved along two lines. 
\textbf{2D video synthesis} methods, such as Wav2Lip~\cite{prajwal2020lip}, FREAK~\cite{freak2025}, and audio-centric coupling approaches~\cite{fu2025dual}, can achieve strong lip-sync in constrained settings but lack the 3D geometric consistency required for view-agnostic applications. 
In parallel, \textbf{3D Morphable Model (3DMM)} based approaches~\cite{blanz1999morphable} offer explicit geometric control, yet their low-dimensional parameterization often produces over-smoothed, ``puppet-like'' faces with limited realism. 
Recent diffusion-based portrait animation and video generation approaches (e.g., Hallo3~\cite{hallo3}, Ditto~\cite{ditto}), alongside structured perceptual preference optimization for visual generation~\cite{ni2026seeingwm}, further improve 2D visual fidelity, but do not provide explicit, stable 3D geometry for consistent avatar animation and view control.

\subsection{Neural Rendering for Talking Heads: NeRF and 3DGS}
Neural Radiance Fields (NeRF)~\cite{mildenhall2020nerf}, first brought to this task by AD-NeRF~\cite{guo2021adnerf}, enabled photorealistic, 3D-consistent talking heads. More recently, 3D Gaussian Splatting (3DGS)~\cite{kerbl20233d} has further improved efficiency. Audio-driven person-specific systems such as GaussianTalker~\cite{gaussiantalker}, TalkingGaussian~\cite{li2024talkinggaussian}, and GSTalker~\cite{gstalker} demonstrate the promise of Gaussian-based representations for talking-face synthesis. In parallel, Gaussian avatar systems such as GaussianSpeech~\cite{gaussianspeech} and AudioRTA~\cite{audiorta} explore broader settings including multi-view capture and telepresence, while GaussianHead-style methods~\cite{gaussianhead} focus more on controllable head-avatar modeling than on monocular audio-driven talking-head synthesis. Despite these advances, most NeRF- and 3DGS-based talkers still rely on \emph{continuous} audio representations as the dominant driving signal. They do not explicitly encode discrete closure-level constraints, making sharp articulations difficult under ambiguity and co-articulation.

\subsection{Linguistic Guidance and 3D Facial Motion Models}
Foundational principles from speech science emphasize that discrete linguistic units should be primary drivers for articulation~\cite{leanderson1971electromyographic, lucero2012speech}. This idea has also been validated in learning-based animation systems: VOCA~\cite{cudeiro2019capture} used phonetic labels to drive a 3DMM and achieved improved lip-sync accuracy, while related systems have explored phonemes as additional supervision or regularization to improve synchronization~\cite{imitator}. A parallel line of work learns speech-driven facial motion on 3D meshes, often with transformers or learned latent codes, as in FaceFormer and CodeTalker~\cite{faceformer, codetalker}. These approaches provide strong geometric control but are typically coupled with mesh rendering or require additional modules to achieve photorealistic appearance. Our work bridges these directions by introducing time-aligned phoneme guidance into a 3DGS avatar pipeline, aiming to improve articulation while preserving the rendering quality and efficiency of Gaussian splatting.

\begin{figure*}[t]
    \centering
    \includegraphics[width=0.95\linewidth]{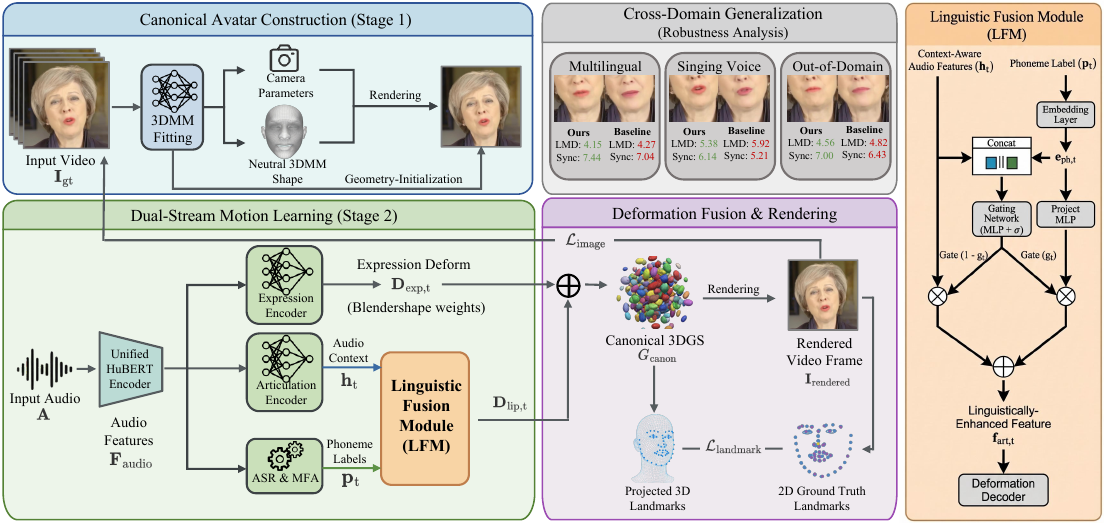} 
   \caption{\textbf{Overview of our Phoneme-Driven Gaussian Splatting (PD-GS) framework.}
\textbf{Stage 1: Canonical avatar construction.} From the training video \(\mathbf{I}_{\text{gt}}\), we fit a 3DMM to recover camera parameters and a neutral facial shape, initializing the canonical 3D Gaussian avatar \(G_{\text{canon}}\).
\textbf{Stage 2: Dual-stream motion learning.} A unified HuBERT encoder extracts audio features \(\mathbf{F}_{\text{audio}}\) from \(\mathbf{A}\), shared by an \emph{expression} stream (blendshape deformation \(\mathbf{D}_{\text{exp},t}\)) and an \emph{articulation} stream (context features \(\mathbf{h}_t\)).
Phoneme labels \(\mathbf{p}_t\) from ASR+MFA are fused with \(\mathbf{h}_t\) via the LFM to predict lip deformation \(\mathbf{D}_{\text{lip},t}\).
The total deformation \(\mathbf{D}_{\text{total},t}=\mathbf{D}_{\text{lip},t}+\mathbf{D}_{\text{exp},t}\) renders \(G_{\text{canon}}\).
Training uses image reconstruction \(\mathcal{L}_{\text{image}}\) and landmark consistency \(\mathcal{L}_{\text{landmark}}\) losses.
We further analyze cross-domain robustness.}
    \Description{The full PD-GS pipeline in two stages. The first stage builds
    a canonical Gaussian avatar from a training video and fitted 3D face
    geometry. The second stage maps audio to expression deformation while a
    parallel ASR and phoneme-alignment path guides lip deformation through the
    linguistic fusion module. Their sum deforms the avatar, and image and lip
    landmark losses supervise training.}

    \label{fig:pipeline}
\end{figure*}

\section{Methodology}

\subsection{Notation and Problem Formulation}
\label{sec:notation}
We denote the training video as a sequence of RGB frames \(\{\mathbf{I}^{t}_{\text{gt}}\}_{t=1}^{T}\) with corresponding audio \(\mathbf{A}\).
We denote the per-frame camera parameters estimated by 3DMM fitting as \(\{\Pi_t\}_{t=1}^{T}\) (intrinsics and extrinsics), which are used for rendering and for projecting 3D landmarks to 2D.
Given a driving audio \(\mathbf{A}\), the goal is to synthesize \(\{\mathbf{I}^{t}_{\text{rendered}}\}_{t=1}^{T}\) for a target identity with photorealistic appearance and linguistically correct articulation.

We extract frame-aligned audio features \(\mathbf{F}_{\text{audio}}=\{\mathbf{f}_t\}_{t=1}^{T}\) using HuBERT~\cite{hsu2021hubert}, and obtain time-aligned phoneme labels \(\mathbf{p}=\{p_t\}_{t=1}^{T}\) using an automated ASR+forced-alignment pipeline~\cite{radford2023robust, mcauliffe2017montreal}.
We use \(G\) to denote a set of 3D Gaussians, and reserve \(g_t\) (or \(\mathbf{g}_t\)) for the gate value in the Linguistic Fusion Module (LFM).

\subsection{Preliminaries: 3D Gaussian Splatting}
\label{sec:preliminaries}
3D Gaussian Splatting (3DGS)~\cite{kerbl20233d} represents a 3D scene using a set of explicit Gaussian primitives \(\{G_i\}_{i=1}^{N}\).
Each Gaussian \(G_i\) is parameterized by \(\theta_i = \{\boldsymbol{\mu}_i, \mathbf{q}_i, \mathbf{s}_i, \mathbf{c}_i, \alpha_i\}\), including its 3D mean \(\boldsymbol{\mu}_i \in \mathbb{R}^3\), rotation (as a quaternion) \(\mathbf{q}_i \in \mathbb{R}^4\), scale \(\mathbf{s}_i \in \mathbb{R}^3\), color represented by spherical harmonics coefficients \(\mathbf{c}_i \in \mathbb{R}^{k \times 3}\), and opacity \(\alpha_i \in \mathbb{R}\).
Given a camera view, Gaussians are projected onto the image plane and composited via differentiable alpha blending to render \(\mathbf{I}_{\text{rendered}}\).

\begin{table*}[t]
    \centering
    \caption{Quantitative comparison with representative talking-head synthesis methods.
    We report reconstruction fidelity (PSNR), perceptual quality (LPIPS, NIQE, BRISQUE), lip-sync geometry (LMD), audio-visual synchronization (Sync), and model size.
    Best results are in \textbf{bold}; second-best results are underlined.
    Results for diffusion-based portrait animation methods marked with $\dagger$ are included for reference, as their original training data and/or evaluation protocols are not fully aligned with the person-specific HDTF setting.}
    
    \label{tab:sota_comparison}
    \normalsize
    \setlength{\tabcolsep}{4pt}
    {%
    \begin{tabular}{@{}l|l|cccccc|c@{}}
        \toprule
        \textbf{Category} & \textbf{Method}
        & \textbf{PSNR} \( \uparrow \)
        & \textbf{LPIPS} \( \downarrow \)
        & \textbf{LMD} \( \downarrow \)
        & \textbf{Sync} \( \uparrow \)
        & \textbf{NIQE} \( \downarrow \)
        & \textbf{BRISQUE} \( \downarrow \)
        & \textbf{Model Size} \( \downarrow \) \\
        \midrule

        \multirow{7}{*}{2D-based}
        & Wav2Lip~\cite{prajwal2020lip} & 27.12 & 0.068 & 5.96 & \textbf{9.13} & 5.53 & 35.11 & 50 MB \\
        & DINet~\cite{dinet}           & 27.91 & 0.051 & 4.81 & 6.72           & 4.87 & 31.73 & 900 MB \\
        & IP-LAP~\cite{iplap}          & 30.14 & 0.049 & 3.93 & 4.92           & 4.36 & 27.90 & 450 MB \\
        & SadTalker~\cite{zhang2023sadtalker} & 32.58 & 0.055 & 4.51 & 8.51 & 4.45 & 28.15 & 1500 MB \\
        & MuseTalk~\cite{musetalk} & 30.92 & 0.047 & 3.82 & 7.11 & 4.28 & 27.42 & 1800 MB \\
        & Hallo3$^\dagger$~\cite{hallo3} & 31.48 & 0.043 & 3.55 & 7.63 & 4.07 & 26.31 & 2100 MB \\
        & Ditto$^\dagger$~\cite{ditto} & 32.04 & 0.039 & 3.31 & 8.02 & 3.94 & 24.88 & 2300 MB \\

        \midrule

        \multirow{6}{*}{NeRF-based}
        & AD-NeRF~\cite{guo2021adnerf} & 27.13 & 0.152 & 3.02 & 4.69 & 6.05 & 40.23 & 30 MB \\
        & RAD-NeRF~\cite{radnerf}      & 31.85 & 0.069 & 2.98 & 4.93 & 4.23 & 26.34 & 28 MB \\
        & ER-NeRF~\cite{ernerf}        & 33.01 & 0.031 & 3.01 & 5.11 & 3.82 & 23.11 & 50 MB \\
        & GeneFace~\cite{geneface}     & 33.22 & \underline{0.029} & \underline{2.70} & 7.84 & \underline{3.71} & \textbf{20.14} & 25 MB \\
        & MimicTalk~\cite{mimictalk}   & 33.09 & 0.031 & 2.81 & 7.92 & 3.81 & 21.52 & 70 MB \\
        & SyncTalk~\cite{synctalk}     & 33.15 & 0.030 & 2.85 & 7.95 & 3.74 & \underline{20.62} & 800 MB \\

        \midrule

        \multirow{3}{*}{3DGS-based}
        & GaussianTalker~\cite{gaussiantalker} & \textbf{33.61} & 0.031 & 2.71 & \underline{8.89} & 3.90 & 22.89 & \textbf{22 MB} \\
        & TalkingGaussian~\cite{li2024talkinggaussian} & 33.45 & 0.033 & 2.97 & 8.67 & 3.95 & 23.50 & 30 MB \\
        & GSTalker~\cite{gstalker} & 33.07 & 0.035 & 2.83 & 5.17 & 3.89 & 21.55 & 30 MB \\
        \cmidrule{2-9}
        & \textbf{PD-GS (Ours)} & \underline{33.55} & \textbf{0.027} & \textbf{2.66} & 8.85 & \textbf{3.61} & 20.77 & \underline{24 MB} \\

        \bottomrule
    \end{tabular}
    }
\end{table*}

\subsection{PD-GS: A Phoneme-Driven Framework}
\label{sec:overview}
As illustrated in Figure~\ref{fig:pipeline}, PD-GS follows a two-stage design:
(1) constructing a canonical avatar \(G_{\text{canon}}\), and
(2) learning a linguistically-enhanced motion generator.

\subsection{Stage 1: Canonical 3DGS Avatar Construction}
\label{sec:canonical_construction}
A canonical avatar is required for realistic animation.
In this stage, our goal is to build a canonical 3DGS model \(G_{\text{canon}}\) that captures the subject’s appearance while providing structured geometry suitable for subsequent deformations.

\paragraph{3DMM-based facial initialization and camera estimation.}
We fit a 3D Morphable Model (3DMM) to the training video to recover per-frame camera parameters (and head pose) and a stable identity shape prior, following common practices in 3D-aware talking head systems~\cite{guo2021adnerf, geneface, gaussiantalker}.
By averaging the identity-related shape parameters across frames and setting expression-related coefficients to neutral, we obtain a stable neutral 3DMM mesh.
We use its vertex positions to initialize a dense set of Gaussian means \(\boldsymbol{\mu}\) in the facial region, which provides a complete and topologically valid facial structure and improves optimization stability.

\paragraph{Non-facial regions: hair, shoulders, and background.}
Since the 3DMM prior mainly covers the inner face, we additionally represent non-facial regions (hair, shoulders, and background) with Gaussians as part of \(G_{\text{canon}}\).
In our implementation, these non-facial Gaussians are initialized by \textbf{randomly sampling 3D points} in a loose capture volume around the head/upper-body region, and are then refined via the standard 3DGS optimization with \textbf{adaptive densification}~\cite{kerbl20233d}.
This strategy avoids relying on SfM point clouds and keeps the pipeline fully automatic while allowing 3DGS to allocate capacity to visually complex non-facial areas during densification.

\paragraph{Optimization and adaptive densification.}
Starting from the above initialization, we optimize \(\theta = \{\boldsymbol{\mu}, \mathbf{q}, \mathbf{s}, \mathbf{c}, \alpha\}\) by rendering \(G_{\text{canon}}\) under the recovered camera parameters and minimizing an image reconstruction loss \(\mathcal{L}_{\text{image}}\) (Section~\ref{sec:training}) against \(\mathbf{I}^{t}_{\text{gt}}\).
We follow the standard 3DGS procedure~\cite{kerbl20233d}, including \textbf{adaptive densification} that periodically clones/splits Gaussians to allocate capacity to visually complex regions (e.g., eyes and lips).

\subsection{Stage 2: Dual-Stream Motion Generator}
\label{sec:motion_generator}
The motion generator drives \(G_{\text{canon}}\) with audio (and phoneme) inputs.
It adopts a decoupled design with two parallel streams that predict distinct components of motion:
(i) high-frequency lip/jaw articulation and
(ii) low-frequency expression and head-related dynamics.
This separation lets each stream specialize temporally.

\paragraph{Unified audio representation.}
We employ \textbf{HuBERT}~\cite{hsu2021hubert} to extract a frame-level audio feature sequence \(\mathbf{F}_{\text{audio}} \in \mathbb{R}^{T \times D}\) from \(\mathbf{A}\).
We temporally align these features with the video frame rate so that each time step \(t\) has a corresponding audio feature vector \(\mathbf{f}_t\).
The resulting \(\mathbf{F}_{\text{audio}}\) is shared by both motion streams.

\paragraph{Articulation module with Linguistic Fusion Module (LFM).}
The articulation module is designed to generate precise lip and jaw motion by resolving the ambiguity of purely audio-driven prediction.
It consists of (i) an articulation encoder that produces a context-aware audio feature \(\mathbf{h}_t\), and (ii) the proposed \textbf{Linguistic Fusion Module (LFM)} that injects time-aligned phoneme cues.

\textbf{Phoneme extraction, vocabulary, and frame alignment.}
We obtain phoneme intervals from an automatic ASR+forced-alignment pipeline. Let
\(\{(s_k,e_k,\phi_k)\}_{k=1}^{K}\) denote the aligned phoneme intervals, where \(s_k\) and \(e_k\) are the start/end times and \(\phi_k\) is the phoneme label. We build a phoneme vocabulary of size \(V=40\), including silence and pause tokens. Each frame-level phoneme token is mapped to a learnable embedding \(\mathbf{e}_{\text{ph},t}\in\mathbb{R}^{64}\). To align phonemes to video frames, we assign frame \(t\), whose center timestamp is \(\tau_t\), to the phoneme interval satisfying \(\tau_t\in[s_k,e_k)\). Therefore, a phoneme may span multiple consecutive frames. For boundary frames, we assign the phoneme whose interval contains the frame center (equivalently, the one with larger temporal overlap in our implementation).

\textbf{Gated fusion.}
Given the audio context feature \(\mathbf{h}_t\in\mathbb{R}^{D}\) and the phoneme embedding \(\mathbf{e}_{\text{ph},t}\), the LFM predicts a dynamic gate vector:
\begin{equation}
\mathbf{g}_t = \sigma\left(\text{MLP}_{\text{gate}}([\mathbf{h}_t, \mathbf{e}_{\text{ph}, t}])\right),
\end{equation}
where \(\mathbf{g}_t \in [0,1]^D\) is a \emph{vector gate} with the same dimensionality as \(\mathbf{h}_t\), enabling channel-wise modulation in the latent articulation space. We project the phoneme embedding to the same feature space with \(\text{MLP}_{\text{ph}}\), and obtain the linguistically-enhanced feature:
\begin{equation}
\mathbf{f}_{\text{art}, t} = (\mathbf{1} - \mathbf{g}_t) \odot \mathbf{h}_t
+ \mathbf{g}_t \odot \text{MLP}_{\text{ph}}(\mathbf{e}_{\text{ph}, t}),
\label{eq:lfm}
\end{equation}
where \(\odot\) denotes element-wise multiplication. Because the gate is predicted jointly from audio and phoneme features, it adaptively adjusts the relative contribution of the two modalities according to their learned compatibility. Individual channels of \(\mathbf{g}_t\) do not have predefined phonetic meanings; for interpretability, we visualize the mean gate activation \(\bar g_t = \frac{1}{D}\sum_{j=1}^{D} g_{t,j}\).

\textbf{Lip deformation prediction.}
A decoder MLP predicts the articulatory deformation \(\mathbf{D}_{\text{lip}, t}\) from \(\mathbf{f}_{\text{art}, t}\).
(Here \(\mathbf{D}_{\text{lip}, t}\) denotes the predicted deformation applied to the avatar geometry; see the next paragraph for how it is used to deform 3DGS.)

\paragraph{Expression module (including eye blink / upper-face motion / head-related motion).}
In parallel, the expression module generates non-articulatory motions such as eyebrow movement, blinks, and low-frequency head dynamics.
It is driven by \(\mathbf{F}_{\text{audio}}\) using a dedicated temporal encoder, enabling specialization: the articulation stream focuses on high-frequency phonetic transitions, while the expression stream models lower-frequency prosodic dynamics~\cite{busso2008emotion}.
To promote stability, the module predicts weights over a compact expression blendshape basis~\cite{cao2014facewarehouse}, from which the expression deformation \(\mathbf{D}_{\text{exp}, t}\) is computed.

\paragraph{Deformation fusion and applying motion to 3DGS.}
We fuse the predicted articulatory deformation \(\mathbf{D}_{\text{lip}, t}\) and expressive deformation \(\mathbf{D}_{\text{exp}, t}\) to obtain the total deformation \(\mathbf{D}_{\text{total}, t}\).
This deformation is applied to the canonical avatar \(G_{\text{canon}}\) to produce the animated Gaussians for rendering at time \(t\).

Unlike variants that freeze appearance parameters during motion learning, in our final setting we \textbf{update the full 3DGS parameter set} \(\theta=\{\boldsymbol{\mu}, \mathbf{q}, \mathbf{s}, \mathbf{c}, \alpha\}\) during Stage 2 training together with the motion generator.
This end-to-end optimization allows the canonical representation to adapt to the subsequent non-rigid deformations and improves overall stability and fidelity.
At inference time, the learned \(G_{\text{canon}}\) parameters are fixed, and the motion is realized by applying the predicted deformations to generate the time-varying rendered frames.

\subsection{Training Strategy and Objective}
\label{sec:training}
Following prior 3D-aware avatar works~\cite{gaussiantalker, geneface}, we adopt a two-stage training strategy to decouple identity modeling from motion learning and improve optimization stability.

\paragraph{Stage 1: Canonical avatar training.}
We first learn the static canonical avatar \(G_{\text{canon}}\) by optimizing \(\{\boldsymbol{\mu}, \mathbf{q}, \mathbf{s}, \mathbf{c}, \alpha\}\) (Section~\ref{sec:canonical_construction}).
The objective is to minimize the image reconstruction loss \(\mathcal{L}_{\text{image}}\) between rendered frames and ground-truth frames.

\paragraph{Stage 2: Motion Generator Training.}
We train the Dual-Stream Motion Generator end-to-end.
In this stage, we jointly optimize the motion generator and the 3DGS parameters of \(G_{\text{canon}}\), i.e., \(\{\boldsymbol{\mu}, \mathbf{q}, \mathbf{s}, \mathbf{c}, \alpha\}\), under the image-domain supervision described below.

\paragraph{Image reconstruction loss.}
We use a combination of an L1 loss and LPIPS~\cite{zhang2018unreasonable}:
\begin{align}
    \mathcal{L}_{\text{image}} ={}& \lambda_{\text{L1}} \lVert\mathbf{I}^{t}_{\text{rendered}} - \mathbf{I}^{t}_{\text{gt}}\rVert_1 \notag \\
    & + \lambda_{\text{LPIPS}} \mathcal{L}_{\text{LPIPS}}(\mathbf{I}^{t}_{\text{rendered}}, \mathbf{I}^{t}_{\text{gt}}).
\end{align}

\paragraph{Landmark consistency loss.}
To provide an explicit geometric constraint for lip motion, we employ a landmark consistency loss \(\mathcal{L}_{\text{landmark}}\), and identify corresponding 3D vertices on a canonical proxy mesh, apply the predicted lip deformation \(\mathbf{D}_{\text{lip}, t}\), and project them to 2D using \(\Pi_t\) to obtain \(\mathbf{L}_{\text{proj}}^{2D}\); Specifically, we use a pre-trained 2D landmark detector to extract lip landmarks \(\mathbf{L}^{2D}_{\text{gt}}\) from \(\mathbf{I}^{t}_{\text{gt}}\), and compute projected landmarks \(\mathbf{L}^{2D}_{\text{proj}}\) from the deformed 3D geometry using known camera parameters.
We minimize:
\begin{equation}
    \mathcal{L}_{\text{landmark}} = \lVert\mathbf{L}^{2D}_{\text{proj}} - \mathbf{L}^{2D}_{\text{gt}}\rVert_2^2.
\end{equation}
The final loss is:
\begin{equation}
    \mathcal{L}_{\text{total}} = \mathcal{L}_{\text{image}} + \lambda_{\text{landmark}} \mathcal{L}_{\text{landmark}}.
\end{equation}

\begin{figure*}[t]
    \centering
    \includegraphics[width=\linewidth]{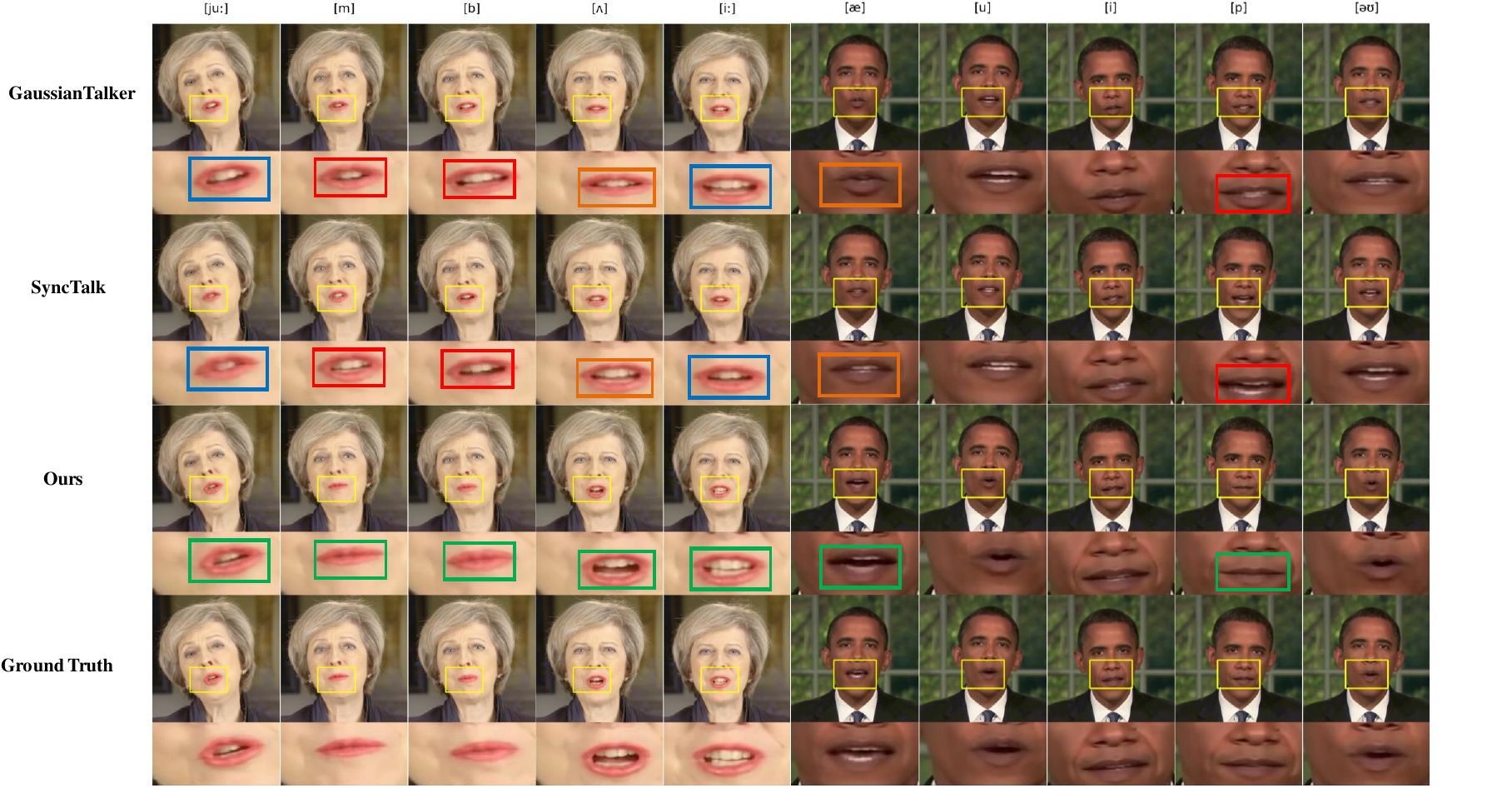} 
    \caption{\textbf{Qualitative analysis of articulation errors on challenging phonemes.}
    We highlight common failure modes of audio-driven baselines:
    \emph{closure failure} (incomplete lip sealing for bilabial stops, e.g., [p], producing ``leaky mouth''),
    \emph{phonetic ambiguity} (distinct vowels, e.g., [\textipa{\ae}] and [\textipa{3}], rendered with similarly open mouths), and
    \emph{weak rounding} (insufficient lip protrusion, e.g., [ju:]).
    In contrast, \emph{ours} produces clearer closures and more distinctive articulations via explicit phoneme guidance.
    Colors are used only as a visual aid.}
    \Description{Rows of talking-head frames compare several baselines with
    PD-GS for difficult phonemes. The examples mark incomplete bilabial
    closure, similar mouth shapes for different vowels, and insufficient lip
    rounding. The PD-GS row shows closed lips or more distinctive rounded and
    open shapes in the corresponding cases.}
    \label{fig:phoneme_comparison}
\end{figure*}

\section{Experiments}
\label{sec:experiments}

We conduct comprehensive experiments to evaluate the effectiveness of PD-GS and to answer the following questions:
(1) How does PD-GS compare with strong recent talking head synthesis methods in visual quality and lip-sync accuracy?
(2) What is the contribution of the Linguistic Fusion Module (LFM)?
We provide quantitative and qualitative evidence demonstrating the benefits of explicit phoneme guidance.

\subsection{Experimental Setup}

\paragraph{Datasets.}
We follow standard protocols on the HDTF dataset~\cite{zhang2021flow}, consistent with prior person-specific talking head works such as AD-NeRF~\cite{guo2021adnerf} and GaussianTalker~\cite{gaussiantalker}.
Unless otherwise specified, we report results on two widely used subjects (``Obama'' and ``May'') to enable direct comparison with existing 3D-aware baselines.
Videos are at 25 fps.
Following common practice, the ``Obama'' sequence is processed at \(450 \times 450\), while other subjects are processed at \(512 \times 512\), with the speaker centered in the frame.

\paragraph{Baselines.}
We compare PD-GS with representative methods spanning five paradigms.

\textbf{(i) 2D portrait talkers.}
We include strong 2D-based lip-sync and portrait reenactment baselines that are widely used in practice, including Wav2Lip~\cite{prajwal2020lip}, DINet~\cite{dinet}, and IP-LAP~\cite{iplap}.
We also include recent high-quality portrait talkers that integrate 3D priors or strong image generators, such as SadTalker~\cite{zhang2023sadtalker} and MuseTalk~\cite{musetalk}.

\textbf{(ii) Diffusion-based portrait animation.}
To reflect recent progress, we additionally discuss diffusion-based and motion-diffusion talkers, such as Hallo3~\cite{hallo3} and Ditto~\cite{ditto}.
These methods are typically designed for generalizable, single-image portrait animation and may use different training data and evaluation protocols.
Therefore, we include them primarily as reference methods in the quantitative table and for qualitative/efficiency discussion, rather than as strictly protocol-aligned person-specific baselines.

\textbf{(iii) 3D-aware radiance-field talkers.}
We include representative NeRF-based approaches, including AD-NeRF~\cite{guo2021adnerf}, RAD-NeRF~\cite{radnerf}, ER-NeRF~\cite{ernerf}, GeneFace~\cite{geneface}, SyncTalk~\cite{synctalk}, and MimicTalk~\cite{mimictalk}.

\textbf{(iv) 3DGS-based talkers.}
To directly compare within the same rendering paradigm, we evaluate leading 3DGS-based talking-face methods, including GaussianTalker~\cite{gaussiantalker}, TalkingGaussian~\cite{li2024talkinggaussian}, and, when available under an aligned person-specific protocol, GSTalker~\cite{gstalker}. We mention GaussianHead-style methods~\cite{gaussianhead} in Related Work as relevant 3DGS head-avatar modeling approaches, but do not treat them as directly comparable baselines because they are not monocular audio-driven talking-head synthesis methods under the same evaluation setting.

\textbf{(v) Multi-view / real-time avatar systems.}
We also discuss recent multi-view Gaussian avatar systems and real-time telepresence systems, e.g., GaussianSpeech~\cite{gaussianspeech} and AudioRTA~\cite{audiorta}.
These methods require multi-view capture or a different data/setting, hence they are not directly comparable to our monocular HDTF protocol; we report them for context and efficiency reference.

\paragraph{Implementation Details.}
Our model is implemented in PyTorch.
We train the canonical avatar for 30k iterations (Stage 1), and then train the motion generator for 100k iterations (Stage 2) on a single NVIDIA RTX 4090 GPU.
During Stage 2, the motion generator and the parameters of the canonical 3DGS avatar are jointly optimized. 
This allows the Gaussian representation to adapt to the predicted non-rigid deformations while preserving stable identity and texture.
For baseline methods, we use officially released code and follow their recommended configurations on HDTF to ensure fair comparison.
For person-specific methods (e.g., ours, GeneFace, RAD-NeRF, GaussianTalker, and TalkingGaussian), we train a separate model per subject.
For methods designed for generalization (e.g., Wav2Lip and SadTalker), we use their provided pre-trained weights.

\begin{figure*}[t]
    \centering
    \includegraphics[width=\linewidth]{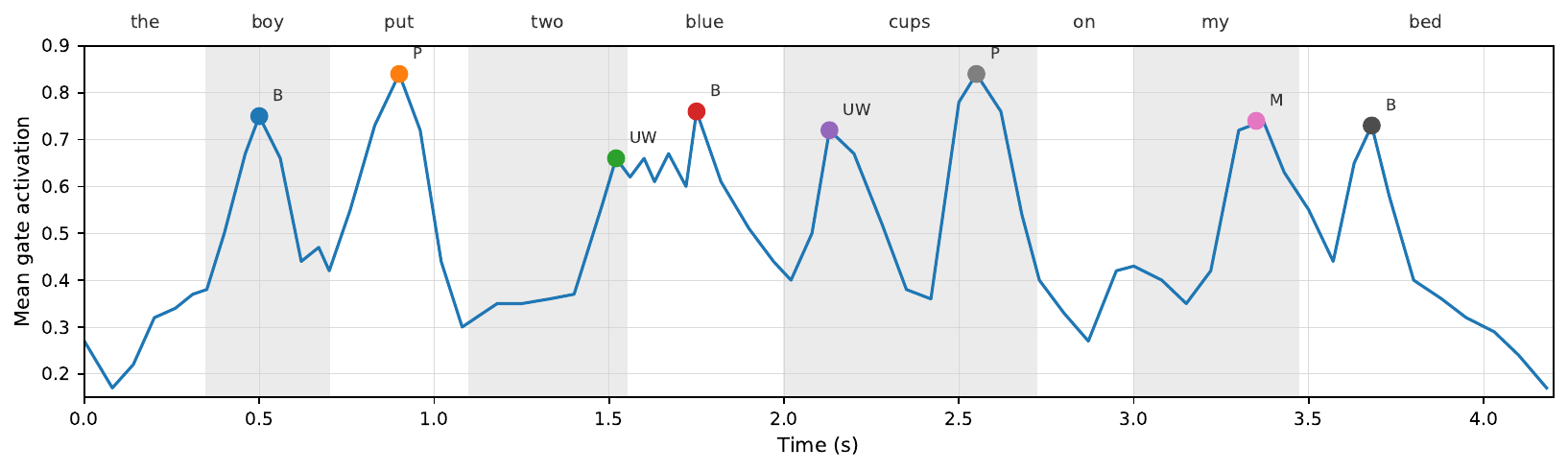}
    \caption{\textbf{Mean gate activation over time on a held-out utterance.} The gate \(\bar g_t=\frac{1}{D}\sum_j g_{t,j}\) varies systematically across aligned phoneme intervals instead of collapsing to a constant weight, showing that the LFM adaptively modulates phoneme cues, with local increases near articulation-critical segments such as bilabial closures and strong rounding.}
    \Description{A time-series plot of mean fusion-gate activation with
    phoneme intervals shown along the horizontal axis. The curve changes across
    phoneme boundaries and has local increases near articulation-critical
    segments instead of remaining constant.}
    \label{fig:gate_curve}
\end{figure*}

\subsection{Quantitative Analysis}

\paragraph{Comparison with Representative Baselines.}
Table~\ref{tab:sota_comparison} summarizes quantitative results on HDTF.
PD-GS achieves the best overall performance on the most articulation-relevant metric, \textbf{Lip Landmark Distance (LMD)}, reaching \textbf{2.66}, the best result among the compared baselines. This improves upon the strongest 3DGS baseline (GaussianTalker, 2.71) and directly supports our central claim: injecting discrete phonetic information produces more geometrically accurate and physically plausible lip motion.

PD-GS also achieves strong perceptual quality, attaining the best scores on \textbf{LPIPS} and \textbf{NIQE}, while remaining highly competitive on \textbf{BRISQUE}. 
Importantly, these gains come with a compact model size of 24 MB, highlighting the efficiency of PD-GS compared to much larger systems (e.g., SadTalker and DINet).

While Wav2Lip attains the highest Sync score, this result should be interpreted together with the nature of the metric: Sync mainly reflects coarse audio-visual temporal consistency, whereas our primary goal is to improve the geometric faithfulness of visible articulation. In particular, incomplete bilabial closures or insufficient vowel distinctiveness may not be fully penalized by a SyncNet-style score even when they are clearly visible. Under this perspective, PD-GS remains highly competitive on Sync (8.85, close to GaussianTalker's 8.89) while achieving the best LMD, indicating a better balance between temporal alignment and physically plausible lip geometry.

Finally, PD-GS maintains strong reconstruction fidelity, with PSNR close to the best-performing 3DGS baseline.
Overall, PD-GS offers a strong balance between articulation precision, perceptual quality, synchronization, and model efficiency.

\paragraph{Evaluation Metrics.}
We evaluate methods using complementary metrics: PSNR measures pixel-level reconstruction fidelity; LPIPS, NIQE, and BRISQUE assess perceptual quality; LMD directly measures geometric discrepancies of lip landmarks; and Sync reflects audio-visual temporal alignment.

Because SyncNet-style scores target coarse synchronization rather than closure-level geometric accuracy, we treat LMD as the primary articulation-sensitive metric and regard Sync as complementary; PD-GS remains in the first tier on Sync while substantially improving the more geometry-sensitive LMD.

\paragraph{Efficiency (FPS and latency).}
PD-GS runs at 110 FPS / 9.1 ms per frame in \emph{model-side} inference, i.e., after HuBERT features and aligned phoneme tokens are available; because the phoneme pipeline uses offline ASR+MFA preprocessing, this reflects low-latency model inference rather than fully streaming end-to-end latency. A full comparison is in the supplementary material.

\paragraph{Ablation Study.}
To validate the contribution of the Linguistic Fusion Module (LFM), we perform ablations summarized in Table~\ref{tab:ablation}:
\begin{itemize}
    \item \textbf{w/o LFM}: Remove the entire linguistic fusion pathway, reducing the articulation module to an audio-only predictor.
    \item \textbf{w/o Gating}: Replace the learned gating mechanism with simple feature concatenation, to test whether dynamic fusion is necessary.
\end{itemize}
Removing LFM significantly degrades both LMD and Sync, confirming that phoneme information is critical for high-precision articulation.
Moreover, the performance drop with ``w/o Gating'' demonstrates that dynamic gated fusion is more effective than naive concatenation.

\begin{figure}[t]
    \centering
    \includegraphics[width=0.95\columnwidth]{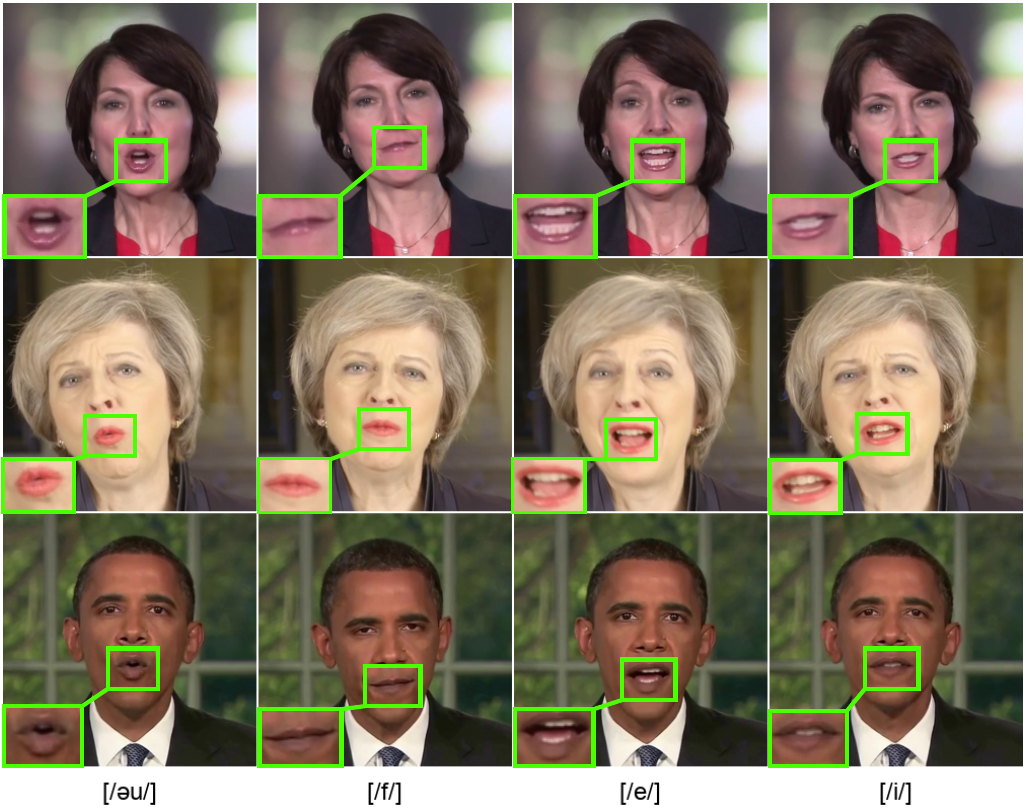} 
    \caption{\textbf{Phoneme consistency across identities.} For the same phoneme (columns), PD-GS generates consistent and physically correct articulations across three subjects (rows), while naturally adapting to each individual's facial geometry.}
    \Description{A grid of rendered faces with three identities in rows and
    phonemes in columns. Within each column, the identities exhibit the same
    characteristic mouth articulation while retaining their individual facial
    appearance.}
    \label{fig:identity_consistency}
\end{figure}

\begin{table}[t]
    \centering
    \caption{Ablation study on key components. The results highlight the critical role of the Linguistic Fusion Module (LFM) in improving lip-sync accuracy.}
    \label{tab:ablation}
    \footnotesize
    \begin{tabular}{@{}lccc@{}}
        \toprule
        \textbf{Model} & \textbf{LMD} \( \downarrow \) & \textbf{Sync} \( \uparrow \) & \textbf{PSNR} \( \uparrow \) \\
        \midrule
        w/o LFM (Audio-Only) & 2.73 & 8.71 & \textbf{33.60} \\
        w/o Gating (Concat Fusion) & 2.70 & 8.78 & 33.48 \\
        \textbf{Ours (PD-GS, Full)} & \textbf{2.66} & \textbf{8.85} & 33.55 \\
        \bottomrule
    \end{tabular}
\end{table}

While the above metrics evaluate objective reconstruction and lip-sync accuracy, it is also important to assess perceptual quality from a human perspective. Therefore, we conduct a user study using Mean Opinion Score (MOS).

\paragraph{What does the gate learn?}
Figure~\ref{fig:gate_curve} visualizes the temporal gate trajectory on held-out utterances: rather than remaining constant, the gate varies systematically across phoneme segments, indicating that the LFM does not collapse to static fusion.

\subsection{User Study (MOS)}
To complement objective metrics, we conduct a Mean Opinion Score (MOS) study on three perceptual dimensions: \emph{Lip-sync Accuracy}, \emph{Visual Realism}, and \emph{Naturalness}. We randomly sample 24 test clips from the test set, anonymize method names, and invite 18 raters to score each clip on a 1--5 Likert scale (1: very poor, 5: excellent). All clips are presented in randomized order. Table~\ref{tab:mos} reports the average scores across raters. As shown, PD-GS receives the highest MOS on all three dimensions, indicating that explicit phoneme guidance improves not only geometric lip accuracy but also the overall perceptual quality and naturalness of the generated talking heads.

\begin{table}[t]
    \centering
    \caption{Mean Opinion Score (MOS) comparison on subjective quality. Scores are on a 1--5 scale; higher is better.}
    \label{tab:mos}
    \footnotesize
    \setlength{\tabcolsep}{4pt}
    \begin{tabular}{@{}lccc@{}}
        \toprule
        Method & Lip-sync Accuracy $\uparrow$ & Visual Realism $\uparrow$ & Naturalness $\uparrow$ \\
        \midrule
        Wav2Lip & 3.68 & 3.14 & 3.21 \\
        SadTalker & 3.39 & 3.31 & 3.36 \\
        MuseTalk & 3.77 & 3.70 & 3.64 \\
        SyncTalk & 3.89 & 3.75 & 3.79 \\
        GaussianTalker & 3.98 & 3.91 & 3.88 \\
        TalkingGaussian & 3.84 & 3.82 & 3.80 \\
        \textbf{PD-GS (Ours)} & \textbf{4.12} & \textbf{4.01} & \textbf{4.07} \\
        \bottomrule
    \end{tabular}
\end{table}

\subsection{Additional-Dataset Validation}
To verify that the gains of PD-GS are not specific to HDTF, we additionally evaluate on a VoxCeleb2 subset under the same evaluation protocol. Table~\ref{tab:extra_dataset} shows that PD-GS generalizes well to the additional dataset, achieving the best LPIPS, LMD, and NIQE among the compared methods, while remaining competitive in PSNR and Sync. Consistent with the HDTF results, these gains are most evident on the articulation-sensitive metric LMD, indicating that phoneme grounding improves lip geometry beyond a single recording condition.

\begin{table}[t]
    \centering
    \caption{Additional-dataset results on a VoxCeleb2 subset. We compare PD-GS with representative talking-face baselines under the same evaluation protocol. Best results are in \textbf{bold}; second-best results are underlined.}
    \label{tab:extra_dataset}
    \footnotesize
    \setlength{\tabcolsep}{3.5pt}
    \begin{tabular}{@{}lccccc@{}}
        \toprule
        Method & PSNR $\uparrow$ & LPIPS $\downarrow$ & LMD $\downarrow$ & Sync $\uparrow$ & NIQE $\downarrow$ \\
        \midrule
        \multicolumn{6}{c}{Portrait / 2D-based methods} \\
        \midrule
        Wav2Lip & 31.24 & 0.051 & 4.87 & 8.29 & 4.26 \\
        SadTalker & 31.96 & 0.044 & 3.74 & 8.11 & 4.03 \\
        MuseTalk & 32.28 & 0.040 & 3.42 & 8.24 & 3.91 \\
        \midrule
        \multicolumn{6}{c}{3D-aware / Gaussian-based methods} \\
        \midrule
        GeneFace & 32.63 & 0.038 & 3.14 & 8.05 & 3.89 \\
        SyncTalk & 32.68 & 0.035 & 3.10 & 8.22 & \underline{3.84} \\
        GaussianTalker & \textbf{32.94} & 0.036 & 3.08 & \textbf{8.41} & 4.01 \\
        TalkingGaussian & 32.71 & 0.039 & 3.26 & 8.18 & 4.08 \\
        GSTalker & 32.79 & \underline{0.034} & \underline{3.01} & 8.27 & 3.96 \\
        \textbf{PD-GS (Ours)} & \underline{32.88} & \textbf{0.031} & \textbf{2.94} & \underline{8.36} & \textbf{3.79} \\
        \bottomrule
    \end{tabular}
\end{table}

\subsection{Qualitative Analysis}

Figure~\ref{fig:phoneme_comparison} analyzes challenging phonetic sequences: purely audio-driven baselines often produce ambiguous or physically incorrect ``averaged'' mouth configurations, whereas PD-GS resolves closure failures and improves phonetic distinctiveness via the LFM.

Figure~\ref{fig:identity_consistency} further shows that this phoneme-to-articulation mapping generalizes across identities, producing consistent articulations for the same phoneme while adapting to each speaker's facial shape.

\section{Conclusion}
\label{sec:conclusion}
We introduced PD-GS to address the persistent ``muffled'' articulation in end-to-end audio-driven talking heads by integrating discrete phonemes with continuous audio representations through a dynamic fusion mechanism, tackling the key limitation of mapping continuous audio to inherently discrete articulatory events without explicit linguistic structure.

Extensive experiments show that PD-GS produces sharper, more accurate, and more robust lip motion while maintaining photorealistic rendering quality and compact model size; future work may explore few-shot generalization and additional articulators such as the tongue.

By combining speech-driven linguistic cues with modern neural rendering, PD-GS improves articulation fidelity in photorealistic talking-head synthesis.

\clearpage

\begin{acks}
This work is supported in part by the National Natural Science Foundation
of China under Grant 62476054, and the Fundamental Research Funds for the
Central Universities of China. This research work is supported by the Big
Data Computing Center of Southeast University.
\end{acks}

\bibliographystyle{ACM-Reference-Format}
\bibliography{sample-base}

@String{Computing = "Computing" }

@String{Computer = "{IEEE} Computer" }

@String{Springer = "Springer-Verlag" }

@String(CVPR= {IEEE Conf. Comput. Vis. Pattern Recog.})

@String(ICCV= {Int. Conf. Comput. Vis.})

@String(AAAI = {AAAI})

@article{leanderson1971electromyographic,
   title={Electromyographic studies of facial muscle activity in speech},
   author={Leanderson, R and Persson, A and {\"O}hman, S},
   journal={Acta oto-laryngologica},
   volume={72},
   number={1-6},
   pages={361--369},
   year={1971},
   publisher={Taylor \& Francis}
 }

@inproceedings{blanz1999morphable,
  author    = {Blanz, V. and Vetter, T.},
  title     = {A Morphable Model for the Synthesis of 3D Faces},
  booktitle = {Proceedings of the 26th Annual Conference on Computer Graphics and Interactive Techniques ({SIGGRAPH})},
  year      = {1999}
}

@inproceedings{busso2008emotion,
  title={Analysis of emotion recognition using facial expressions, speech and multimodal information},
  author={Busso, Carlos and Deng, Zhigang and Yildirim, Serdar and Bulut, Murtaza and Lee, Chul Min and Kazemzadeh, Abe and Lee, Sungbok and Neumann, Ulrich and Narayanan, Shrikanth},
  booktitle={Proceedings of the 6th international conference on Multimodal interfaces},
  pages={205--211},
  year={2004}
}

@article{cao2014facewarehouse,
  title={Facewarehouse: A 3d facial expression database for visual computing},
  author={Cao, Chen and Weng, Yanlin and Zhou, Shun and Tong, Yiying and Zhou, Kun},
  journal={IEEE Transactions on Visualization and Computer Graphics},
  volume={20},
  number={3},
  pages={413--425},
  year={2013},
  publisher={IEEE}
}

@inproceedings{cudeiro2019capture,
  author    = {Cudeiro, D. and Bolkart, T. and Laidlaw, C. and Ranjan, A. and Black, M. J.},
  title     = {Capture, Learning, and Synthesis of 3D Speaking Styles},
  booktitle = {Proceedings of the IEEE/CVF Conference on Computer Vision and Pattern Recognition},
  year      = {2019}
}

@inproceedings{guo2021adnerf,
  title={Ad-nerf: Audio driven neural radiance fields for talking head synthesis},
  author={Guo, Yudong and Chen, Keyu and Liang, Sen and Liu, Yong-Jin and Bao, Hujun and Zhang, Juyong},
  booktitle={Proceedings of the IEEE/CVF international conference on computer vision},
  pages={5784--5794},
  year={2021}
}

@article{kerbl20233d,
  title={3D Gaussian splatting for real-time radiance field rendering.},
  author={Kerbl, Bernhard and Kopanas, Georgios and Leimk{\"u}hler, Thomas and Drettakis, George},
  journal={ACM Trans. Graph.},
  volume={42},
  number={4},
  pages={139--1},
  year={2023}
}

@article{lucero2012speech,
title = {Speech production: Models and data},
journal = {Speech Communication},
volume = {22},
number = {2},
pages = {89-92},
year = {1997},
issn = {0167-6393},
doi = {https://doi.org/10.1016/S0167-6393(97)00028-9},
url = {https://www.sciencedirect.com/science/article/pii/S0167639397000289},
author = {Pascal Perrier and Rafael Laboissière and Christian Abry and Shinji Maeda}
}

@inproceedings{mcauliffe2017montreal,
  title={Montreal forced aligner: Trainable text-speech alignment using kaldi.},
  author={McAuliffe, Michael and Socolof, Michaela and Mihuc, Sarah and Wagner, Michael and Sonderegger, Morgan},
  booktitle={Interspeech},
  volume={2017},
  pages={498--502},
  year={2017}
}

@article{mildenhall2020nerf,
  title={Nerf: Representing scenes as neural radiance fields for view synthesis},
  author={Mildenhall, Ben and Srinivasan, Pratul P and Tancik, Matthew and Barron, Jonathan T and Ramamoorthi, Ravi and Ng, Ren},
  journal={Communications of the ACM},
  volume={65},
  number={1},
  pages={99--106},
  year={2021},
  publisher={ACM New York, NY, USA}
}

@inproceedings{prajwal2020lip,
  title={A lip sync expert is all you need for speech to lip generation in the wild},
  author={Prajwal, KR and Mukhopadhyay, Rudrabha and Namboodiri, Vinay P and Jawahar, CV},
  booktitle={Proceedings of the 28th ACM international conference on multimedia},
  pages={484--492},
  year={2020}
}

@inproceedings{zhang2018unreasonable,
  title={The unreasonable effectiveness of deep features as a perceptual metric},
  author={Zhang, Richard and Isola, Phillip and Efros, Alexei A and Shechtman, Eli and Wang, Oliver},
  booktitle={Proceedings of the IEEE conference on computer vision and pattern recognition},
  pages={586--595},
  year={2018}
}

@article{hsu2021hubert,
  title={Hubert: Self-supervised speech representation learning by masked prediction of hidden units},
  author={Hsu, Wei-Ning and Bolte, Benjamin and Tsai, Yao-Hung Hubert and Lakhotia, Kushal and Salakhutdinov, Ruslan and Mohamed, Abdelrahman},
  journal={IEEE/ACM transactions on audio, speech, and language processing},
  volume={29},
  pages={3451--3460},
  year={2021},
  publisher={IEEE}
}

@inproceedings{radford2023robust,
  title={Robust speech recognition via large-scale weak supervision},
  author={Radford, Alec and Kim, Jong Wook and Xu, Tao and Brockman, Greg and McLeavey, Christine and Sutskever, Ilya},
  booktitle={International conference on machine learning},
  pages={28492--28518},
  year={2023},
  organization={PMLR}
}

@inproceedings{zhang2023sadtalker,
  title={Sadtalker: Learning realistic 3d motion coefficients for stylized audio-driven single image talking face animation},
  author={Zhang, Wenxuan and Cun, Xiaodong and Wang, Xuan and Zhang, Yong and Shen, Xi and Guo, Yu and Shan, Ying and Wang, Fei},
  booktitle={Proceedings of the IEEE/CVF conference on computer vision and pattern recognition},
  pages={8652--8661},
  year={2023}
}

@inproceedings{zhang2021flow,
  title={Flow-guided one-shot talking face generation with a high-resolution audio-visual dataset},
  author={Zhang, Zhimeng and Li, Lincheng and Ding, Yu and Fan, Changjie},
  booktitle={Proceedings of the IEEE/CVF conference on computer vision and pattern recognition},
  pages={3661--3670},
  year={2021}
}

@inproceedings{dinet,
  author    = {Zhang, Z. and Hu, Z. and Deng, W. and Fan, C. and Lv, T. and Ding, Y.},
  title     = {{DINet}: Deformation Inpainting Network for Realistic Face Visually Dubbing on High Resolution Video},
  booktitle = {Proceedings of the AAAI Conference on Artificial Intelligence},
  year      = {2023}
}

@inproceedings{iplap,
  author    = {Zhong, Weizhi and Fang, Chaowei and Cai, Yinqi and Wei, Pengxu and Zhao, Gangming and Lin, Liang and Li, Guanbin},
  title     = {Identity-preserving Talking Face Generation with Landmark and Appearance Priors},
  booktitle = {Proceedings of the IEEE/CVF conference on computer vision and pattern recognition},
  year      = {2023}
}

@misc{musetalk,
  author       = {Yang, Zhitong and Wang, Zhong and Liu, Shiguang and Yu, Butian and Yan, Xin and Shao, Hao},
  title        = {MuseTalk: A Full-Body Video-Driven Talking Head Framework with Expressive Speech Styles},
  year         = {2024},
  howpublished = {arXiv preprint arXiv:2401.06820}
}

@inproceedings{ernerf,
  author    = {Li, Jiahe and Zhang, Jiawei and Bai, Xiao and Zhou, Jun and Gu, Lin},
  title     = {Efficient Region-Aware Neural Radiance Fields for High-Fidelity Talking Portrait Synthesis},
  booktitle = {Proceedings of the IEEE/CVF International Conference on Computer Vision},
  year      = {2023}
}

@inproceedings{mimictalk,
  author    = {Gao, Peng and Song, Sicheng and Li, Chen and Liu, Yang and Liu, Zichao and Xu, Ying},
  title     = {MimicTalk: Generalizable Talking Head Synthesis in the Wild},
  booktitle = {Proceedings of the IEEE/CVF Conference on Computer Vision and Pattern Recognition},
  year      = {2024}
}

@inproceedings{synctalk,
  author    = {Peng, Ziqiao and Hu, Wentao and Shi, Yue and Zhu, Xiangyu and Zhang, Xiaomei and Zhao, Hao and He, Jun and Liu, Hongyan and Fan, Zhaoxin},
  title     = {{SyncTalk}: The Devil Is in the Synchronization for Talking Head Synthesis},
  booktitle = {Proceedings of the IEEE/CVF Conference on Computer Vision and Pattern Recognition},
  year      = {2024}
}

@inproceedings{gaussiantalker,
  author    = {Cho, K. and Lee, J. and Yoon, H. and Hong, Y. and Ko, J. and Ahn, S. and Kim, S.},
  title     = {{GaussianTalker}: Real-time talking head synthesis with 3D Gaussian Splatting},
  booktitle = {Proceedings of the 32nd ACM International Conference on Multimedia},
  year      = {2024}
}

@inproceedings{geneface,
  author    = {Ye, Z. and Jiang, Z. and Ren, Y. and Liu, J. and He, J. and Zhao, Z.},
  title     = {{GeneFace}: Generalized and High-Fidelity Audio-Driven 3D Talking Face Synthesis},
  booktitle = {Proceedings of the IEEE/CVF Conference on Computer Vision and Pattern Recognition},
  year      = {2023}
}

@inproceedings{radnerf,
  author    = {Tang, J. and Wang, K. and Zhou, H. and Chen, X. and He, D. and Hu, T. and Liu, J. and Zeng, G. and Wang, J.},
  title     = {Real-Time Neural Radiance Talking Portrait Synthesis via Audio-Spatial Decomposition},
  booktitle = {European Conference on Computer Vision},
  year      = {2022}
}

@inproceedings{li2024talkinggaussian,
  title={Talkinggaussian: Structure-persistent 3d talking head synthesis via gaussian splatting},
  author={Li, Jiahe and Zhang, Jiawei and Bai, Xiao and Zheng, Jin and Ning, Xin and Zhou, Jun and Gu, Lin},
  booktitle={European Conference on Computer Vision},
  pages={127--145},
  year={2024},
  organization={Springer}
}

@inproceedings{hallo3,
  title     = {Hallo3: Highly Dynamic and Realistic Portrait Image Animation with Video Diffusion Transformer},
  author    = {Cui, Jiahao and Li, Hui and Zhan, Yun and Shang, Hanlin and Cheng, Kaihui and Ma, Yuqi and Mu, Shan and Zhou, Hang and Wang, Jingdong and Zhu, Siyu},
  booktitle = {Proceedings of the IEEE/CVF Conference on Computer Vision and Pattern Recognition (CVPR)},
  pages     = {21086--21095},
  month     = jun,
  year      = {2025}
}

@article{ditto,
  title   = {Ditto: Motion-Space Diffusion for Controllable Realtime Talking Head Synthesis},
  author  = {Li, Tianqi and Zheng, Ruobing and Yang, Minghui and Chen, Jingdong and Yang, Ming},
  journal = {arXiv preprint arXiv:2411.19509},
  year    = {2024}
}

@article{audiorta,
  title   = {Audio Driven Real-Time Facial Animation for Social Telepresence},
  author  = {Lee, Jiye and Li, Chenghui and Tran, Linh and Wei, Shih-En and Saragih, Jason and Richard, Alexander and Joo, Hanbyul and Bai, Shaojie},
  journal = {arXiv preprint arXiv:2510.01176},
  year    = {2025}
}

@article{gaussianspeech,
  title   = {GaussianSpeech: Audio-Driven Gaussian Avatars},
  author  = {Aneja, Shivangi and Sevastopolsky, Artem and Kirschstein, Tobias and Thies, Justus and Dai, Angela and Nie{\ss}ner, Matthias},
  journal = {arXiv preprint arXiv:2411.18675},
  year    = {2024}
}

@inproceedings{faceformer,
  title     = {FaceFormer: Speech-Driven {3D} Facial Animation with Transformers},
  author    = {Fan, Yingruo and Lin, Zhaojiang and Saito, Jun and Wang, Wenping and Komura, Taku},
  booktitle = {Proceedings of the IEEE/CVF Conference on Computer Vision and Pattern Recognition (CVPR)},
  year      = {2022}
}

@article{gstalker,
  title={GSTalker: Real-time Audio-Driven Talking Face Generation via Deformable Gaussian Splatting},
  author={Chen, Bo and Hu, Shoukang and Chen, Qi and Du, Chenpeng and Yi, Ran and Qian, Yanmin and Chen, Xie},
  journal={arXiv preprint arXiv:2404.19040},
  year={2024}
}

@article{gaussianhead,
  title={GaussianHead: High-fidelity Head Avatars with Learnable Gaussian Derivation},
  author={Wang, Jie and Xie, Jiu-Cheng and Li, Xianyan and Xu, Feng and Pun, Chi-Man and Gao, Hao},
  journal={arXiv preprint arXiv:2312.01632},
  year={2023}
}

@inproceedings{codetalker,
  title     = {CodeTalker: Speech-Driven {3D} Facial Animation with Discrete Motion Prior},
  author    = {Xing, Jinbo and Xia, Menghan and Zhang, Yuechen and Cun, Xiaodong and Wang, Jue and Wong, Tien-Tsin},
  booktitle = {Proceedings of the IEEE/CVF Conference on Computer Vision and Pattern Recognition (CVPR)},
  year      = {2023}
}

@inproceedings{imitator,
  title     = {Imitator: Personalized Speech-driven {3D} Facial Animation},
  author    = {Thambiraja, Balamurugan and Habibie, Ikhsanul and Aliakbarian, Sadegh and Cosker, Darren and Theobalt, Christian and Thies, Justus},
  booktitle = {Proceedings of the IEEE/CVF International Conference on Computer Vision (ICCV)},
  year      = {2023}
}

@inproceedings{fu2025dual,
  author    = {Fu, Ao and Ni, Ziqi and Zhou, Yi},
  title     = {Dual Audio-Centric Modality Coupling for Talking Head Generation},
  booktitle = {2025 International Conference on Virtual Reality and Visualization (ICVRV)},
  year      = {2025},
  pages     = {884--889},
  doi       = {10.1109/ICVRV67992.2025.00155}
}

@inproceedings{freak2025,
  author    = {Ni, Ziqi and Fu, Ao and Zhou, Yi},
  title     = {FREAK: Frequency-modulated High-fidelity and Real-time Audio-driven Talking Portrait Synthesis},
  booktitle = {Proceedings of the 2025 International Conference on Multimedia Retrieval (ICMR '25)},
  year      = {2025},
  pages     = {1036--1044},
  publisher = {Association for Computing Machinery},
  address   = {New York, NY, USA},
  url       = {https://doi.org/10.1145/3731715.3733344},
  doi       = {10.1145/3731715.3733344}
}

@inproceedings{ni2026seeingwm,
  author    = {Ni, Ziqi and Liang, Yuanzhi and Li, Rui and Zhou, Yi and Huang, Haibin and Zhang, Chi and Li, Xuelong},
  title     = {Seeing What Matters: Visual Preference Policy Optimization for Visual Generation},
  booktitle = {Proceedings of the IEEE/CVF Conference on Computer Vision and Pattern Recognition (CVPR)},
  month     = {June},
  year      = {2026},
  pages     = {27260--27269}
}

\end{document}